# Incorporating Uncertainty in Learning to Defer Algorithms for Safe Computer-Aided Diagnosis

Jessie Liu[1,2], Blanca Gallego[1], Sebastiano Barbieri[1]
[1]Centre for Big Data Research in Health, University of New South Wales

## Abstract

Deep neural networks are increasingly being used for computer-aided diagnosis, but erroneous diagnoses can be extremely costly for patients. We propose a learning to defer with uncertainty (LDU) algorithm which identifies patients for whom diagnostic uncertainty is high and defers them for evaluation by human experts. LDU was evaluated on the diagnosis of myocardial infarction (using discharge summaries), the diagnosis of any comorbidities (using structured data), and the diagnosis of pleural effusion and pneumothorax (using chest x-rays), and compared with 'learning to defer without uncertainty information' (LD) and 'direct triage by uncertainty' (DT) methods. LDU achieved the same F1 score as LD but deferred considerably fewer patients (e.g. 36% vs. 69% deferral rate for diagnosing pleural effusion with an F1 score of 0.96). Furthermore, even when many patients were assigned the wrong diagnosis with high confidence (e.g. for the diagnosis of any comorbidities) LDU achieved a 17% increase in F1 score, whereas DT was not applicable. Importantly, the weight of the defer loss in LDU can be easily adjusted to obtain the desired trade-off between diagnostic accuracy and deferral rate. In conclusion, LDU can readily augment any existing diagnostic network to reduce the risk of erroneous diagnoses in clinical practice.

## Introduction

Neural networks with multiple hidden layers (deep learning algorithms) are increasingly being applied to electronic medical records, clinical notes, and medical images for diagnostic purposes. Computer-aided diagnosis has the potential to reduce erroneous decisions and the resource burden imposed on clinical staff. Nevertheless, the use of deep learning for automation in healthcare remains a cause for concern [1], because of the limited interpretability of these methods and because erroneous diagnoses or predictions can be extremely costly for patients [2].

A simple approach to limit the number of erroneous computer-aided diagnoses consists in trusting the model's predictions only if the associated output probabilities are above a specified threshold, but this is problematic due to the poor calibration of deep neural networks [3]. An alternative approach consists of automating the diagnosis of a patient only if the expected model error is lower than the expected human error [2]. The expected model error can be determined by calibrating the output probabilities, or by training an auxiliary neural network to predict the probability of a wrong diagnosis based on the patient's input data. The human error can be estimated by assessing the clinician's accuracy on similar, previously seen, patients, or by building another neural network which predicts the probability of experts' disagreement regarding a patient's diagnosis [4].

---

[2] Corresponding to jessie.liu1@unsw.edu.au

Another recent stream of research, including learning with rejection [5] and learning to defer methods [6][7], proposes to build neural networks which identify groups of patients whose diagnosis can be automated with high accuracy. Learning with rejection [5] simultaneously learns a classifier for the diagnostic task and a rejection function. Such an approach is effective when the optimal rejection region cannot be defined as a simple function of the predicted diagnoses. Learning to defer [6][7] considers the model and human decision makers together to optimize the system's overall accuracy. It does this by automatically diagnosing groups of patients for whom the model is highly accurate and, at the same time, deferring the remaining patients for evaluation by a human expert. Two different approaches have been suggested to combine computer-aided and human diagnoses: one approach starts by training a neural network to predict either the diagnostic class or a binary defer indicator; deferred patients are then passed to human decision makers who might have access to additional patient information or domain expertise [6]. Another approach also trains a neural network to either diagnose or defer, but includes the cost of deferring decisions to human experts within the network's loss function. Specifically, the loss function is the sum of two terms: a classification loss term and a defer loss term [7]. When diagnoses by human experts on the training data are not available, the cost of deferring to experts can be assumed to be constant [7], and the problem becomes to learn a classifier and a defer indicator in one training process.

In this study we propose the Learning to Defer with Uncertainty (LDU) algorithm which considers the diagnostic network's predictive uncertainty when learning which patients to diagnose automatically and which patients to defer to human experts. Our aim is to minimize patients' risk when machine learning (ML) models are deployed in healthcare settings, by preventing the application of computer-aided diagnoses in groups of patients for whom the expected diagnostic error is large. The hypothesis to be tested in this study is that LDU results in higher diagnostic accuracy and fewer deferred patients when compared with learning to defer (without uncertainty) and direct triage by uncertainty algorithms, across different types of data including electronic medical records, clinical notes, and x-ray images.

## Methods

### LDU algorithm

The LDU algorithm consists of two stages (illustrated in Figure 1): in stage one an ensemble of neural networks is trained for the diagnostic task, and diagnoses are determined for every patient with associated uncertainty measures; in stage two a 'learning to defer' neural network takes as input the predicted diagnoses and uncertainty measures from stage one and outputs either the patient's diagnosis or the decision to defer to a human expert.

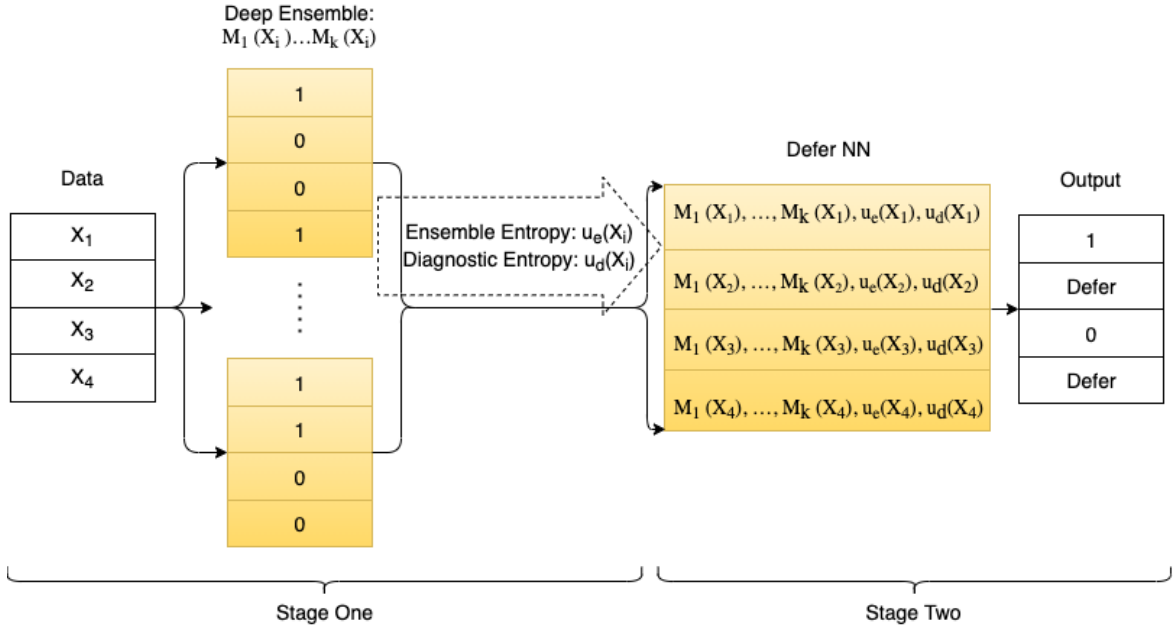

**Figure 1.** Learning to defer using ensemble predictions and their entropy.

There are two sources of uncertainty for a ML model: aleatoric uncertainty (caused by noise inherent to the observed data) and epistemic uncertainty (associated with the distribution of the model's parameter values). Both sources of uncertainty impact the probability distribution of predicted diagnoses; however, aleatoric uncertainty is generally more difficult to quantify because of the lack of multiple measurements of the same variable. Deep ensembles, i.e. ensembles of deep neural networks trained with different random initializations [9], have been found sufficient for capturing epistemic uncertainty, without the need for additional bootstrapping of the training data.

In this study, epistemic uncertainty is quantified by computing both the entropy of the continuous probabilities computed by a deep ensemble (denoted as 'ensemble entropy') and the entropy of the expected probabilities for each diagnosis (denoted as 'diagnostic entropy') [8].

More formally, the ensemble entropy is calculated as:
$$u_e(x_i) = -\sum_{k=1}^{K} P_k(x_i) \log P_k(x_i) \quad (I)$$

where K is the number of deep neural networks in the ensemble and $P_k(x_i)$ is the probability of a positive diagnosis computed by the k-th neural network based on patient data $x_i$.

Similarly, the diagnostic entropy is calculated as:
$$u_d(x_i) = -\sum_{c=1}^{C} P_c(x_i) \log P_c(x_i) \quad (II)$$

where C is the number of classes (C=2 for binary classification) and $P_c(x_i)$ is the fraction of neural networks in the ensemble that predict that input $x_i$ belongs to class c. The more evenly the deep ensemble predictions are distributed between positive and negative classes, the larger the diagnostic entropy will be, which represents higher uncertainty.

In stage two, a learning to defer neural network is built. The network takes as input the output probabilities of the deep ensemble ($M_1(x_i)$, $M_2(x_i)$,,..., $M_k(x_i)$) and the associated ensemble entropy and diagnostic entropy, for a total of K+2 inputs. It outputs either the predicted diagnostic class or the defer class: $f(M_1(x_i), M_2(x_i),,..., M_k(x_i), u(x_i), u_b(x_i)) = \text{class}_{c \in (1,...,C)}$ or defer

The learning to defer network was trained using a previously described loss function [7]. This loss function includes two terms: a cross-entropy loss with the *target* class ($c \in 1, ..., C$), and a weighted cross-entropy loss with the *defer* class. With j $\in$ ($\text{class}_1, ..., \text{class}_C$, defer), the loss can be described by Equation (III). The parameter α is used to adjust the weight of samples that the model decides to defer: by decreasing α, the network is encouraged to defer patients for human evaluation, and vice versa.

$$\text{loss}(x, \text{target}, \text{defer}) = -\log\left(\frac{\exp(x[\text{target}])}{\sum_j \exp(x[j])}\right) - \alpha \log\left(\frac{\exp(x[\text{defer}])}{\sum_j \exp(x[j])}\right) \quad \text{(III)}$$

The proposed LDU algorithm was evaluated on three diagnostic tasks: (1) diagnosis of myocardial infarction using free-text discharge summaries from the MIMIC-III database, (2) diagnosis of any comorbidities (positive Charlson Index) using structured hospital records from the Heritage Health dataset (this diagnostic task has been used previously to evaluate learning to defer methods [6]), and (3) diagnosis of pleural effusion and diagnosis of pneumothorax using chest x-ray images from the MIMIC-CXR database.

### Datasets and preprocessing

**Diagnosis of myocardial infarction (MI):** 13,805 discharge summary notes for ICU patients at Beth Israel Deaconess Medical Center in Boston, MA, USA, publicly available in the MIMIC-III database [10], were used to determine myocardial infarction events. Patients were categorized as MI positive if they had at least one diagnosis of myocardial infarction (ICD codes starting with 410) at hospital discharge, or MI negative if they never had any diagnosis of MI. There were 6,183 discharge summaries for patients with MI and 51,093 discharge summaries for the negative group. To address the imbalance between patients with and without MI, the negative group was down sampled to have the same number of hospital admissions (identified by HADM_ID) as the positive group, this resulted in 6,183 discharge summary notes in the positive group, and 7622 discharge summary notes in the negative group. The free text data were pre-processed by removing de-identification expressions and format characters, and tokenized using the bio discharge summary BERT tokenizer [11]. Long discharge summary notes were split into sequences, with 512 token representations in each sequence.

**Diagnosis of any comorbidities**: The Heritage Health dataset, a structured dataset provided by the Heritage Provider Network (HPN) and containing data of members over a 48-month period, including demographic information (age and sex), claim information such as length of hospital stays, primary condition groups, procedure groups, number of drug prescriptions and laboratory tests, and days since first services, was used to predict the presence of any comorbidities (positive Charlson Index). 21,361 members were included in this diagnostic task, and members with a positive Charlson Index for any claim were assigned a positive target label, all others were assigned a negative label. Features were extracted by encoding categorical data as numeric representations and aggregating the data by member ID. This resulted in 16 features for each member. The majority class was down sampled to have the same number of members within each class (N=7,363 for each class).

**Diagnosis of pleural effusion and pneumothorax:** Chest x-ray images from the MIMIC-CXR database were used for two diagnostic tasks: 17,102 images were used for the diagnosis of pleural effusion and 15,838 images were used for the diagnosis of pneumothorax. The dataset provides ground truth diagnoses for the two conditions and metadata including view positions [14]. For both diagnostic tasks, patients were divided into non-overlapping positive and negative groups depending on whether they had at least one positive chest x-ray for the condition of interest. Then, anteroposterior (AP) view images were selected for every patient and the majority class was down sampled. The process resulted in 17,102 images for the pleural effusion dataset, and 15,838 images for the pneumothorax dataset.

For all diagnostic tasks, the data was split into training and testing datasets based on a 70-30 ratio.

All analyses were performed in accordance with the guidelines and regulations stated in the PhysioNet Credentialed Health Data License: https://physionet.org/content/mimiciii/view-license/1.4/ and https://physionet.org/content/mimic-cxr/view-license/1.0.0/.

## Diagnostic neural networks

Three different types of neural networks were developed for each diagnostic task in this study. These diagnostic networks were used in the first stage of the proposed LDU algorithm (as deep ensembles), as well as to develop baseline algorithms for comparison with the LDU algorithm.

### Diagnosis of myocardial infarction
A BERT-based neural network was developed using an LSTM layer on top of bio discharge summary BERT [11].The last output of the LSTM layer was passed through a fully connected layer with two outputs (for binary classification) and a sigmoid activation function [12][13]. Only the LSTM layer and the final fully connected layer were fine-tuned during training. The network was trained over 18 epochs using an ADAM optimizer with a learning rate of 1e-4 and weight decay of 0.

### Diagnosis of any comorbidities
For this task, we developed a fully connected neural network to predict whether a patient's Charlson Index was positive. The model had two hidden layers (each one with 200 nodes and a sigmoid activation function after the first layer) and two softmax-activated outputs. The model was trained over 4 epochs, again using an ADAM optimizer with a learning rate of 9e-4 and weight decay of 0.

### Diagnosis of pleural effusion and pneumothorax
A DenseNet121 model [15][16][17] was used for both diagnostic tasks after transforming and normalizing the input images. Both training processes used an ADAM optimizer with a learning rate of 1e-4 and weight decay of 1e-5. The network was trained over 2 epochs for the diagnosis of pleural effusion, and over 4 epochs for the diagnosis of pneumothorax.

## LDU implementation details

The deep ensembles for stage one of the LDU approach consisted of 50 randomly initialized diagnostic neural networks. The learning to defer neural network in stage two of the LDU approach was fully connected with two hidden layers (100 nodes each). For the myocardial infarction, pleural effusion and pneumothorax diagnostic tasks, the network was trained over 20 epochs using stochastic gradient descent with a learning rate of 5e-5. For the comorbidity prediction task, the network was trained over 20 epochs using an ADAM optimizer with a learning rate of 9e-4.

The implementation of the LDU algorithm for the three diagnostic tasks described in this paper, together with libraries for applying LDU to other diagnostic tasks is available at https://github.com/liu-res/Learning-to-Defer-with-Uncertainty.

## Comparison with existing algorithms

The proposed LDU algorithm was compared with two other triage algorithms: learning to defer (without uncertainty information) and direct triage by uncertainty.

### Leaning to defer without uncertainty information (LD)
Previously proposed learning to defer algorithms do not consider uncertainty information [6][7]. For every task, learning to defer networks were developed by modifying the corresponding diagnostic neural networks: the output size of the network was increased by one, i.e. if the original binary classification outputs were 0 and 1, then the modified outputs were 0, 1 and 2 (defer class). The loss function in Equation (III) was used for training. Adjusting the weight $\alpha$ of the defer loss led to different trade-offs between model performance and defer rates.

### Direct triage by uncertainty (DT)
For each diagnostic task, deep ensembles [9] consisting of 50 randomly initialized diagnostic neural networks were trained. Predictive uncertainty was estimated by calculating the diagnostic entropy for every patient. Patients whose diagnostic entropy was above a specified threshold were deferred to human experts. Adjusting the entropy threshold led to different trade-offs between model performance and defer rates. Diagnostic entropy (entropy of the expected probabilities for each diagnosis) was used instead of ensemble entropy (entropy of the ensemble output probabilities), because of its superior performance (see also supplemental Figure 1).

## Evaluation metrics

We compared the performance between (1) the proposed LDU algorithm, (2) learning to defer without considering predictive uncertainty information [6][7] (denoted as "LD" for simplicity), and (3) direct triage by uncertainty (denoted as "DT" for simplicity). Two types of F1 scores were computed for every model: an F1 score for patients who were not deferred to a human expert (denoted as "F1"); and an overall F1 score for the entire patient group estimated assuming that patients who are deferred to a human expert all receive the correct diagnosis ("F1 Overall"). The F1 score of the diagnostic neural network (without defer option) is reported as well (denoted as "Diagnostic Network F1"). We systematically varied the weight $\alpha$ of the defer loss for methods (1) and (2) and the entropy threshold for method (3) in training to achieve F1 scores for non-deferred patients above the diagnostic network F1 scores, and measured the corresponding defer rates. Furthermore, performance metrics including accuracy, sensitivity, and specificity were compared between LDU and LD for selected arbitrary F1 scores.

# Results

### Diagnosis of myocardial infarction

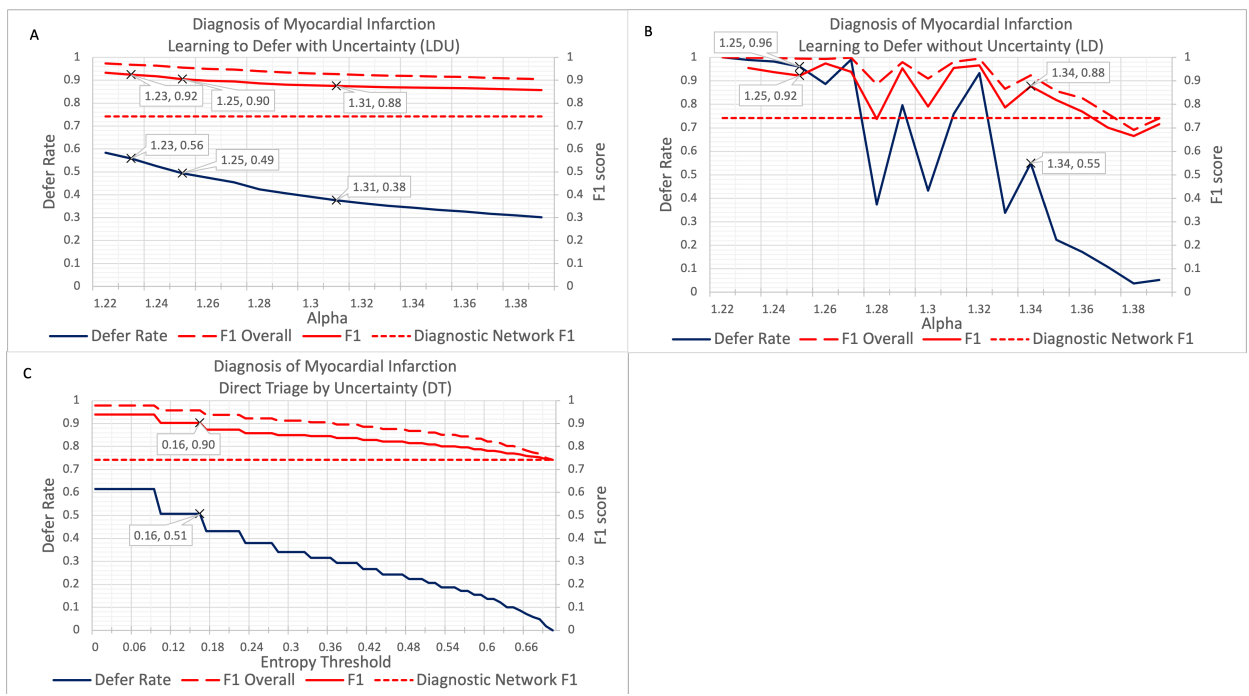

**Figure 2.** Performance comparison for the diagnosis of myocardial infarction using MIMIC-III discharge summaries and the following algorithms: (A) Learning to defer with uncertainty (LDU), (B) Learning to defer without uncertainty information (LD), (C) Direct triage by uncertainty (DT). Each panel shows the F1 scores (red line) for patients who are not deferred to human experts, 'F1 Overall' scores (red dashed line) and the corresponding defer rates (blue line) for different values of the weight of the defer loss α ((A) and (B)) or of the entropy threshold ((C)). The diagnostic network F1 score (without the defer option) is 0.74 (red dotted line).

Figure 2 shows the F1 scores and defer rates for the diagnosis of myocardial infarction from discharge summaries, using three algorithms: LDU, LD and DT. All algorithms were able to achieve a F1 score of 0.90 or more (22% increase over the diagnostic network's F1 score of 0.74) by deferring some patients for human expert evaluation. The LDU algorithm deferred a slightly lower proportion of patients than the DT algorithm (49% vs 51%, respectively). However, the LDU algorithm deferred considerably less patients than the LD algorithm: e.g. to achieve an F1 score of 0.88 (an approximately 20% increase over the diagnostic network's F1 score), the LDU algorithm deferred only 38% of patients while the LD algorithm deferred 55% of patients. In other words, LDU can automatically diagnose a larger portion of patients (therefore requiring less human effort) for the same level of performance increase as the previously proposed LD method.

Importantly, the LDU's F1 score and defer rate increased monotonically as we decreased the weight of the defer loss (α) during training to encourage the algorithm to defer patients to human experts. Specifically, as α was reduced from 1.31 to 1.23, LDU's defer rate increased from 38% to 56%, and meanwhile the F1 score increases monotonically from 0.88 to 0.92. This was not the case for LD: even though it was possible to achieve the same F1 score when α equaled 1.25, the defer rate and F1 score fluctuated as α was decreased.

The DT algorithm showed similar monotonicity and performance as the LDU algorithm. For the DT algorithm, as the uncertainty threshold decreased to 0, the proportion of deferred patients reached its maximum (60%). DT's performance couldn't be improved further. This shows that the proportion of patients whose diagnostic entropy is zero determine the ceiling for performance increase through patient deferral for the DT algorithm.

The monotonicity in LDU's and DT's performance suggests that higher F1 scores can be expected as more patients are deferred. We can determine an optimal trade-off between performance and defer rate by adjusting the parameter α of LDU during training or the entropy threshold of DT. This is not necessarily true for the LD algorithm.

**Diagnosis of any comorbidities**

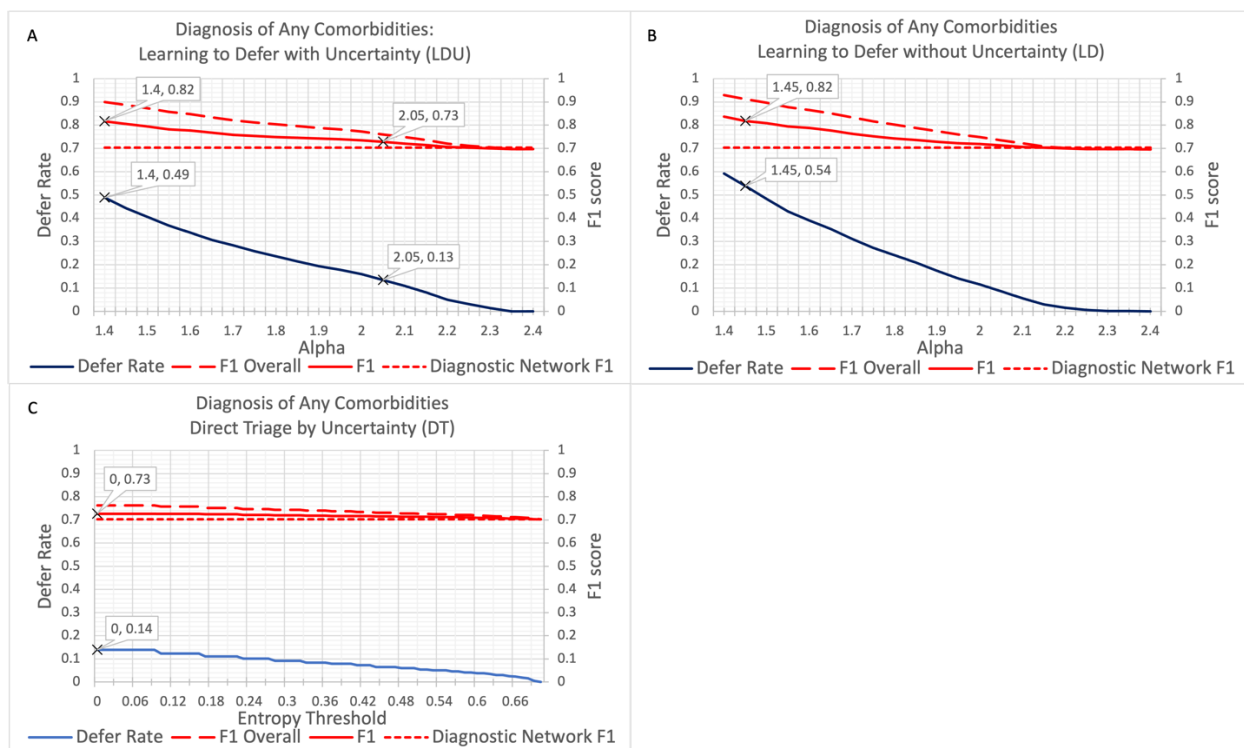

**Figure 3.** Performance comparison for the diagnosis of any comorbidities (identified by positive Charlson Index) using the Heritage Health dataset and the following algorithms: (A) Learning to defer with Uncertainty (LDU), (B) Learning to defer without uncertainty information (LD), (3) Direct triage by uncertainty (DT). Each panel shows the F1 scores (red line) for patients who are not deferred to human experts, 'F1 Overall' scores (red dashed line) and the corresponding defer rates (blue line) for different values of the weight of the defer loss α ((A) and (B)) or of the entropy threshold ((C)). The diagnostic network F1 score (without the defer option) is 0.70 (red dotted line).

Figure 3 shows the performance of LDU algorithm, LD algorithm, and DT algorithm for the diagnosis of comorbidities from structured electronic medical records. The LDU and LD algorithms performed similarly well, with comparable defer rates for specific F1 scores: for the F1 score to reach 0.82 (an approximately 17% increase over the diagnostic network's F1 score of 0.70), the LD algorithm deferred 54% of patients for human evaluation, while the LDU algorithm deferred 49% of patients.

LD's F1 scores and defer rates increased monotonically in this diagnostic task, but not in the previous myocardial infarction diagnosis task. Possible reasons are considered in the Discussion section.

The DT approach provided only a minor benefit for this diagnostic task: because the diagnostic entropy was zero for most patients (86%), the maximum defer rate for the DT algorithm was 14% and the corresponding maximum F1 score was 0.73 (4% increase over the diagnostic network's F1 score). This again illustrates the limited utility of the DT algorithm when the diagnostic entropy is zero for most patients.

These results suggest that even when the architecture of the underlying diagnostic network is relatively simple, the proposed LDU algorithm can lead to considerable gains in F1 score by deferring a smaller proportion of patients than the LD algorithm. Furthermore, the defer rate of the LDU algorithm is less sensitive than the DT algorithm to the distribution of predicted diagnoses (i.e. the diagnostic entropy) for each patient.

**Diagnosis of pleural effusion and pneumothorax**

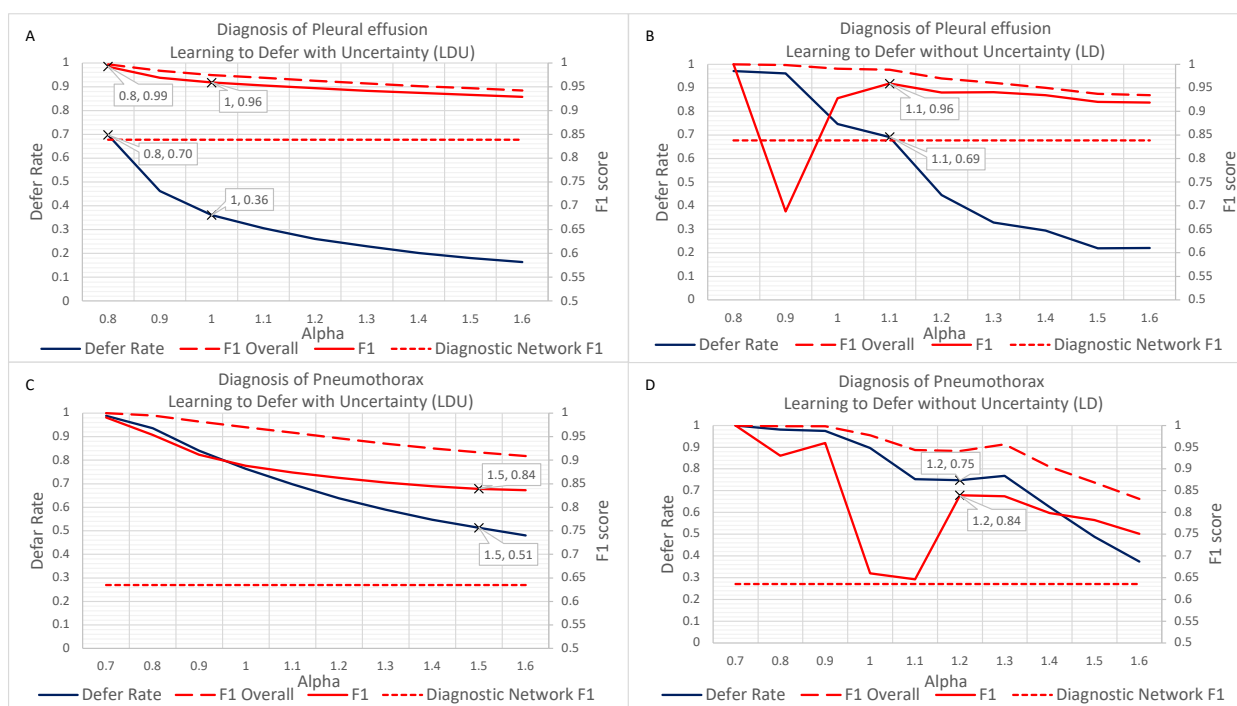

**Figure 4.** Performance comparison for the diagnostic tasks using MIMIC-CXR x-ray images: diagnosis of pleural effusion using (A) learning to defer with uncertainty (LDU) and (B) learning to defer without uncertainty information (LD), diagnosis of pneumothorax using (C) learning to defer with uncertainty (LDU) and (D) learning to defer without uncertainty information (LD). Each panel shows the F1 scores (red line) for patients who are not deferred to human experts, 'F1 Overall' scores (red dashed line) and the corresponding defer rates (blue line) for different values of the weight of the defer loss α. The diagnostic network F1 score (without the defer option) is 0.84 for pleural effusion, and 0.63 for pneumothorax (red dotted line).

Results for the image-based diagnostic tasks support previous findings using free-text and structured tabular data (Figure 4). In particular, the LDU algorithm deferred less patients for human evaluation than the LD algorithm, while reaching the same F1 score. In the diagnosis of pleural effusion, to achieve an F1 score of 0.96 (14% increase over the diagnostic network's F1 score), the LD algorithm deferred 69% of patients while the LDU algorithm deferred only 36% of patients. In the diagnosis of pneumothorax, to achieve an F1 score of 0.84 (a lower target than for pleural effusion because of the more difficult task, but a 25% increase over the diagnostic network's F1 score of 0.63), the LDU algorithm deferred only 51% of patients while the LD algorithm deferred 75% of patients. The LDU algorithm was sufficiently accurate for a considerable proportion of patients even though the performance of the single underlying diagnostic network was poor.

In both diagnostic tasks, the F1 scores of the LDU algorithm increased monotonically as more patients were deferred to human experts.

| α | Defer Rate | F1 | Accuracy | Sensitivity | Specificity |
|---|---|---|---|---|---|
| 0.77 | 81.91% | 0.998 | 0.997 | 0.970 | 1.000 |
| 0.78 | 77.80% | 0.996 | 0.994 | 0.980 | 0.998 |
| 0.79 | 73.64% | 0.993 | 0.991 | 0.983 | 0.995 |
| 0.80 | 69.74% | 0.992 | 0.990 | 0.984 | 0.994 |

**Table 1.** Performance metrics of the LDU algorithm in the diagnosis of pleural effusion, by setting the weight of defer loss between 0.77 and 0.80.

Table 1 shows additional performance metrics for the LDU algorithm when diagnosing pleural effusion and varying the weight α of the defer loss between 0.77 and 0.80. The LDU algorithm identified about 18% to 30% of patients for whom the predicted diagnoses were almost certainly correct. This suggests that the LDU algorithm can identify a subgroup of patients for whom the diagnosis of pleural effusion can be automated with considerably low risk (the error probability can be controlled within 1%).

For both image-based diagnostic tasks, the diagnostic entropies were zero for all patients, i.e. the diagnoses predicted by the deep ensemble had different probabilities but all pointed to the same class. Therefore, the DT algorithm could not be directly applied to these tasks. Using the ensemble entropies instead of the diagnostic entropies did not lead to any increase in F1 scores either (supplemental Figure 1). In comparison, the LDU algorithm could achieve high F1 scores and low defer rates using only the probabilities predicted by the deep ensemble and their entropy.

**Summary of accuracy, sensitivity and specificity**

Table 2 shows the defer rates of the LDU algorithm and the LD algorithm for selected F1 scores, as well as the corresponding accuracy, sensitivity and specificity metrics. The table suggests that the LDU algorithm results in better defer rates than the LD algorithm also when performance metrics other than F1 score are considered.

| Data Source | Diagnostic Task | Method | F1 | Defer Rate | Accuracy | Sensitivity | Specificity |
|---|---|---|---|---|---|---|---|
| MIMIC-III | Myocardial Infarction | LD | 0.88 | 55% | 0.84 | 0.70 | 0.93 |
| MIMIC-III | Myocardial Infarction | LDU | 0.88 | 38% | 0.90 | 0.94 | 0.84 |
| Heritage Health | Comorbidity | LD | 0.82 | 54% | 0.81 | 0.79 | 0.83 |
| Heritage Health | Comorbidity | LDU | 0.82 | 49% | 0.80 | 0.76 | 0.85 |
| MIMIC-CXR | Pleural Effusion | LD | 0.96 | 69% | 0.96 | 0.97 | 0.96 |
| | | LDU | 0.96 | 36% | 0.96 | 0.98 | 0.94 |
| | Pneumothorax | LD | 0.84 | 75% | 0.76 | 0.42 | 0.94 |
| | | LDU | 0.84 | 51% | 0.83 | 0.85 | 0.82 |

**Table 2.** Comparison of accuracy, sensitivity, and specificity between the LDU algorithm and learning to defer without uncertainty (LD) for selected F1 scores, across all diagnostic tasks.

## Discussion

In this study, we propose a novel algorithm (learning to defer with uncertainty, LDU) which considers predictive uncertainty information when identifying patients who should be evaluated by human experts because computer-aided diagnosis is likely to be inaccurate. LDU was found to achieve similar F1 scores, but considerably lower defer rates, than a learning to defer method that does not consider uncertainty information [6][7]. LDU was also associated with monotonic increases in performance and defer rates as the weight of the defer loss was decreased during training, suggesting that an optimal trade-off between LDU's performance and defer rate can easily be found for a variety of diagnostics tasks. In contrast, learning to defer without uncertainty (LD) could not always guarantee increased performance as more patients were being deferred, making its performance in clinical settings somewhat unpredictable.

In one of the diagnostic tasks (diagnosis of any comorbidities), the LD algorithm was associated with monotonic changes in F1 scores and defer rates, similar to the LDU algorithm. We hypothesize that this is due to the relatively small number of parameters of the underlying diagnostic network and consequent low epistemic uncertainty for the predicted diagnoses (also leading to low diagnostic entropy for the LDU and DT algorithms). However, when the diagnostic networks become more complex, such as the BERT-based network in the myocardial infarction diagnostic task, only the LDU algorithm was associated with monotonic improvements in performance as we tuned the parameter $\alpha$.

Furthermore, for tasks such as the diagnosis of pleural effusion, the LDU algorithm was able to identify a subgroup of patients for whom the risk of an erroneous computer-aided diagnosis was close to zero. This might allow the application of computer-aided diagnosis even for diagnostic tasks where the cost of inaccurate predictions is high.

Unlike direct triage by uncertainty [2], LDU performed robustly even in situations where the deep ensemble predicted the wrong diagnosis with high confidence. We hypothesize that individual outlying predictions, which might not be noticeable when applying a simple threshold to the overall diagnostic entropy, are taken into account by LDU's 'learning to defer' neural network.

For the direct triage method, the maximum defer rate occurs at a threshold value of zero, which means that the defer rate can only be as large as the percentage of patients associated with non-zero uncertainty measures. This becomes problematic when a diagnostic network results in uncertainty measures of zero for a large portion of patients, and even more so if the uncertainty is zero for all patients (such as in the diagnosis of pleural effusion and pneumothorax). In these scenarios direct triage by uncertainty is bound to result in poor performance, but the LDU algorithm still led to a wide range of defer rates and large performance gains. Alternatively, we also tried using the ensemble entropy to triage patients, but this did not lead to any increase in performance (as illustrated in supplemental Figure 1), because both correct and incorrect diagnoses can be associated with low entropy as a result of over-confident neural networks.

The LDU algorithm can be easily adapted to a wide range of diagnostic tasks. LDU has advantages when compared to a recent approach that which measures epistemic uncertainty by building fuzzy soft sets of the model's parameters [19], since it does not change any core model parameters, and can be applied to any neural network for classification tasks. Essentially, the LDU algorithm can be used as a "wrapper" to any existing diagnostic model, and the risk of erroneous diagnoses can be reduced by embedding the diagnostic model into the LDU procedure.

Both the LDU algorithm and the direct triage method rely on uncertainty measures to decide which patients should be deferred for human evaluation. Our current approach to capturing epistemic uncertainty is limited: for example, in the diagnosis of pleural effusion and pneumothorax the diagnostic entropy was zero for all patients, possibly due to overconfident predictions by networks in the deep ensemble. Future research could explore whether additional information about epistemic uncertainty could be captured by measuring the entropy of the outputs of intermediate network layers, rather than just of the softmax outputs, and evaluate Bayesian approaches to uncertainty quantification in deep neural networks [18]. Further, this study only considered model uncertainty, and the cost of deferring a patient for human evaluation was assumed to be constant. This constraint is mainly related to the limited availability of public datasets with diagnostic labels from multiple human experts, which could be used to estimate the uncertainty of human diagnoses for each patient. If data with gold standard diagnoses and labels by multiple human experts for each patient were available, the LDU algorithm could be augmented to take into account both the model's and human uncertainty, and patients would be deferred for evaluation by human experts only if the human uncertainty was estimated to be lower than the model's uncertainty.

In conclusion, the proposed LDU algorithm can be used to mitigate the risk of erroneous computer-aided diagnoses in clinical settings. LDU identifies patients for whom the uncertainty of computer-aided diagnosis is estimated to be high and defers them for evaluation by human experts. The algorithm achieves similar diagnostic performance (F1 score, accuracy, sensitivity and specificity) to previous learning to defer algorithms but reduces the proportion of patients deferred for human evaluation. Furthermore, LDU's performance increases monotonically as more patients are deferred, suggesting that the desired trade-off between performance and defer ratio can be obtained for a wide variety of diagnostic tasks.

## Data availability

Applications for access to the MIMIC-III database can be made at https://physionet.org/content/mimiciii/1.4/. The Heritage Health dataset from Kaggle competition (https://www.kaggle.com/c/hhp) can be downloaded at https://foreverdata.org/1015/index.html. Applications for access to the MIMIC-CXR database can be made at https://physionet.org/content/mimic-cxr/2.0.0/

## Code availability

The implementation of the LDU algorithm for the three diagnostic tasks described in this paper, together with libraries for applying LDU to other diagnostic tasks is available at https://github.com/liu-res/Learning-to-Defer-with-Uncertainty.

# Additional information

## Author contributions
J.L. contributed to conceptualisation, data curation, formal analysis, investigation, methodology, software validation, visualisation, writing - original draft.

B.G. contributed to conceptualisation, investigation, project administration, resources, accessing and verifying data, supervision, writing - review & editing.

S.B. contributed to conceptualisation, investigation, methodology, project administration, resources, accessing and verifying data, supervision, writing - review & editing.

## Competing interests
The authors declare no competing interests.